\documentclass[lettersize,journal]{IEEEtran}  %

\usepackage{cite}
\usepackage{float}
\usepackage{multirow}
\usepackage{amssymb,amsfonts}
\usepackage{algorithmic}
\usepackage{booktabs}
\usepackage{graphicx}
\usepackage{textcomp}
\usepackage{xcolor}
\usepackage{algorithm}
\usepackage{verbatim} 
\usepackage{mathrsfs}
\usepackage[caption=false,font=footnotesize]{subfig}
\usepackage{empheq}
\usepackage{lipsum}
\usepackage{comment}
\usepackage{mathtools, cuted}
\usepackage{bm}

\usepackage{amsthm}
\usepackage[utf8]{inputenc}
\usepackage[english]{babel}
\usepackage[colorlinks,linkcolor=black,citecolor=black,urlcolor=black,bookmarks=false,hypertexnames=true]{hyperref} 
\usepackage[textsize=tiny]{todonotes}
\usepackage[font=footnotesize,skip=4pt]{caption}
\usepackage{etoolbox}
\apptocmd{\thebibliography}{\scriptsize}{}{}

\usepackage{soul}

\newtheorem*{definition*}{Definition}



\newtheorem{rem}{Remark}

\begin{document}
\title{STATE-NAV: Stability-Aware Traversability Estimation for Bipedal Navigation on Rough Terrain}

\author{Ziwon Yoon$^{1}$, Lawrence Y. Zhu$^{1}$, Jingxi Lu$^{2}$, Lu Gan$^{1}$, and Ye Zhao$^{1}$%
\vspace{-1\baselineskip}
\thanks{$^{1}$Institute for Robotics and Intelligent Machines, Georgia Institute of Technology, Atlanta, GA 30332, USA 
        {\tt\footnotesize \{zyoon6, lawrencezhu, lgan, yezhao\}@gatech.edu}
        }%
\thanks{$^{2}$Electrical and Computer Engineering, University of Southern California, Los Angeles, CA 90089, USA 
        {\tt\footnotesize jingxil@usc.edu}
        }%
\thanks{This work was supported in part by the Office of Naval Research (ONR) under Grant N000142312223; in part by the National Science Foundation (NSF) under Grant CMMI-2144309, and Grant FRR-2328254; and in part by United States Department of Agriculture (USDA) under Grant 2023-67021-41397.}%
}


\IEEEaftertitletext{\vspace{-6\baselineskip}}
\maketitle
\thispagestyle{empty}
\pagestyle{empty}

\begin{strip}
  \centering
\includegraphics[width=0.96\textwidth]{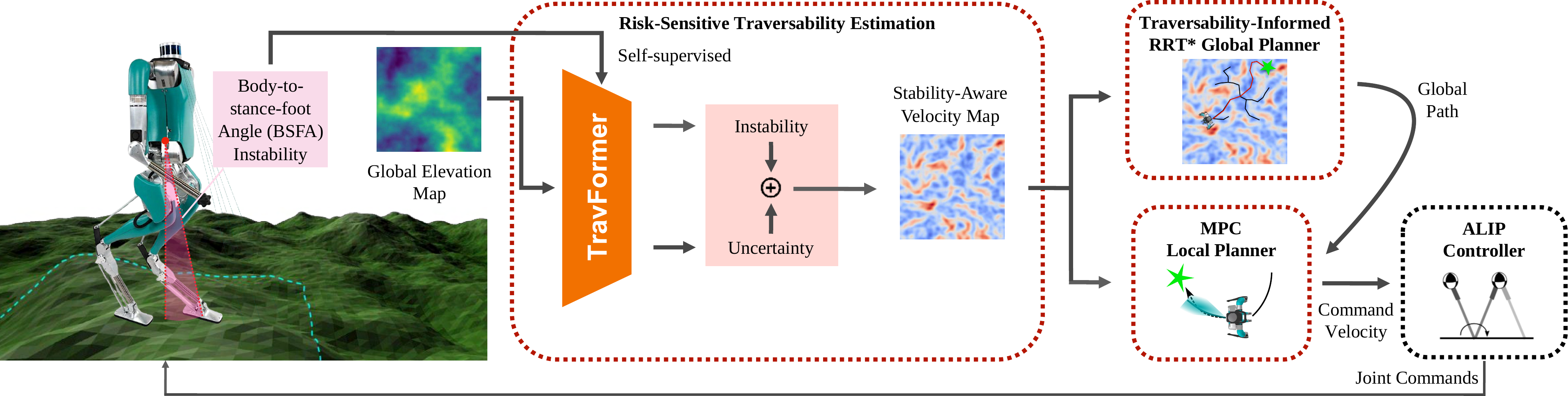}
  \captionof{figure}{Overall diagram of the proposed traversability estimation and the navigation framework. A transformer-based bipedal instability estimator, \textit{TravFormer}, is trained using body-to-stance-foot angle (BSFA) instability as a self-supervised signal to generate stability-aware command velocity map from the geometric representation of the environment. This map is further used by a hierarchically integrated TravRRT* global planner and MPC local planner to safely navigate over diverse rough terrain.}
  \label{fig:overview}
  \vspace{-1\baselineskip}
\end{strip}


\begin{abstract}
Bipedal robots have advantages in maneuvering human-centered environments, but face greater failure risk compared to other stable mobile platforms, such as wheeled or quadrupedal robots. While learning-based traversability has been widely studied for these platforms, bipedal traversability has instead relied on manually designed rules with limited consideration of locomotion stability on rough terrain. In this work, we present the first learning-based traversability estimation and risk-sensitive navigation framework for bipedal robots operating in diverse, uneven environments. TravFormer, a transformer-based neural network, is trained to predict bipedal instability with uncertainty, enabling risk-aware and adaptive planning. Based on the network, we define traversability as stability-aware command velocity—the fastest command velocity that keeps instability below a user-defined limit. This velocity-based traversability is integrated into a hierarchical planner that combines traversability-informed Rapid Random Tree Star (TravRRT*) for time-efficient path planning and Model Predictive Control (MPC) for safe execution. We validate our method in MuJoCo simulator and the real world, demonstrating improved stability, time efficiency, and robustness across diverse terrains compared with existing methods. Project page with code: \href{https://state-nav.github.io/statenav/}{\textcolor{blue}{https://state-nav.github.io/statenav/}}
\end{abstract}

\vspace{-0.1in}

\section{Introduction}\label{sec:introduction}

\begin{figure*}[!t]
  \centering
  \vspace{-0.2in}
\end{figure*}

\IEEEPARstart{W}{ith} the rapid advancement of legged robotics,  humanoid robots have garnered significant attention for their ability to navigate complex environments with rough terrain that are often inaccessible to wheeled mobile systems~\cite{gu2025humanoid, krotkov2018darpa}. 
Despite their advantages, humanoid robots face significant challenges in maintaining stability and balance, particularly on uneven or dynamically changing terrain \cite{huang2023efficient}. Unlike wheeled or quadrupedal robots, which benefit from their more statically stable ground contacts, humanoids have a higher risk of failure due to their limited support area and elevated center of mass, especially during rapid locomotion or abrupt terrain transitions. While various control methods have been proposed to address locomotion stability~\cite{kajita2003biped, gibson2022terrain}, traversability and navigation frameworks that account for the instability of bipedal locomotion on rough terrains remain largely underexplored.

Existing works have explored bipedal traversability using manually designed rules based on geometry~\cite{mccrory2023bipedal, muenprasitivej2024bipedal, shamsah2024terrain}, providing a rough representation of terrain difficulty. However, these manually designed rules often fail to accurately reflect locomotion stability and performance. Meanwhile, for wheeled and quadrupedal robots, extensive research on traversability estimation has leveraged self-supervised learning with various locomotion features such as traction or IMU variances~\cite{wellhausen2019should, wellhausen2021rough, frey2023fast, cai2023probabilistic, cai2024evora, castro2023does, gan2022energy, yang2021real}. However, the relevance of these features to humanoid locomotion stability remains uncertain due to the fundamentally different locomotion strategies employed by bipedal robots. Furthermore, the greater risk of failure in humanoid locomotion requires them to operate under stricter constraints on motion aggressiveness, such as command velocity. This aspect is often overlooked in traditional navigation planning but is crucial for ensuring safe and efficient humanoid navigation.

To address these problems, we take a data-driven approach to evaluate various locomotion features and identify the one that best correlates with bipedal instability for predicting fallover risk. 
A Transformer-based neural network, TravFormer, is then trained to predict the identified metric from terrain features and commanded velocity using simulated data. Instead of directly using the predicted instability as a traversability measure, we define traversability as the \textit{stability-aware command velocity}, i.e., the fastest command velocity that maintains predicted instability below a user-defined threshold, which we use to construct a traversability map over the environment, as illustrated in Fig.~\ref{fig:overview}.

Estimating traversability as a stability-aware command velocity enables stability-prioritized navigation planning. Many traditional planning approaches take traversability as a cost term with tunable weights and jointly optimize it with other costs such as the path length ~\cite{fan2021step, frey2023fast, mccrory2023bipedal, yang2021real, castro2023does, wermelinger2016navigation}. However, varying path length scales across environments often necessitate re-tuning of weights. On the other hand, using stability-aware command velocity mitigates the effort of manual weight tuning, as terrain difficulty is intrinsically encoded in the traversal time via a robot-specific instability threshold instead of environment-specific weights. We leverage this representation in a hierarchical Traversability-Informed Rapidly-exploring Random Tree Star (TravRRT*)--Model Predictive Control (MPC) navigation planning algorithm as shown in Fig. \ref{fig:overview}. The global TravRRT* planner simplifies the planning objective to navigation time optimization, while leveraging terrain traversability to guide informed sampling toward safer, more navigable regions. Additionally, the learned command is incorporated as a constraint in the Linear Inverted Pendulum Model (LIPM)-based local MPC planner, to generate an optimal motion plan that follows the global path while satisfying stability constraints.

In summary, we propose a learning-based bipedal traversability estimation framework that formally incorporates locomotion stability, and design a hierarchical planning framework tailored specifically for humanoid navigation. The key contributions of this paper are summarized as follows:
\begin{itemize}
    \item We conduct a comparative analysis of locomotion features to determine the best feature reflecting bipedal instability--the body-to-stance-foot angle (BSFA)--and use it as the supervisory signal guiding the neural network.
    \item TravFormer is introduced as the first learning-based traversability estimator for humanoid robots on diverse rough terrain, predicting instability with uncertainty using self-supervised labels derived from the identified feature.
    \item Stability-aware bipedal navigation across diverse environments is achieved by representing traversability as a risk-sensitive stability-aware command velocity, providing robust, consistent performance without extensive gain tuning when integrated into the hierarchical TravRRT*–MPC framework. 
    \item We validate the proposed framework in MuJoCo simulator \cite{todorov2012mujoco} and real-world experiments using a humanoid, Digit, demonstrating that our framework enables the robot to navigate diverse rough terrain safely and efficiently.
\end{itemize}

\section{Related Work}\label{sec: Related Works}
\subsection{Self-Supervised Traversability Learning for Grounded Mobile Robots}
Traversability characterizes a robot’s ability to navigate its environment by quantifying terrain difficulty~\cite{papadakis2013terrain}. In existing methods, traversability has been estimated based on manually defined rules based on terrain geometry~\cite{wermelinger2016navigation} or supervised learning with human-labeled semantics~\cite{guan2022ga}. However, both approaches often fail to capture the complex interaction between robots and environments. More recent efforts have developed learning-based traversability estimators in a self-supervised manner, using supervisory signals derived from various proprioceptive features during locomotion, such as ground reaction wrench analysis~\cite{wellhausen2019should}, foothold scores~\cite{wellhausen2021rough}, traction~\cite{frey2023fast, cai2023probabilistic, cai2024evora}, statistical features of sensor measurements such as IMU data~\cite{castro2023does}, and energy consumption metric~\cite{gan2022energy, yang2021real}. While these self-supervising features capture specific aspects of locomotion difficulty of wheeled and quadrupedal robots, they lack validation regarding their ability to represent the fallover risk in case of bipedal locomotion.

\subsection{Traversability Estimation and Navigation Planning for Bipedal Locomotion}
Traversability for bipedal locomotion on rough terrain remains largely underexplored.
In~\cite{mccrory2023bipedal}, A* planning samples candidate footsteps around nodes and defines traversability as the percentage of cells with sufficiently low inclination and height proximity within the sampled region. Other works have explored learning-based traversability estimation for traversing tilted flat surfaces based on footstep availability~\cite{lin2017humanoid,  lin2021long}, or adjusting body height to navigate low-ceiling environments~\cite{li2023autonomous}. These methods target specific settings and do not address large-scale navigation across general rough terrain.
\subsection{Navigation Planning with Traversability Estimation}
Autonomous navigation systems use traversability at multiple levels of the planning stack to navigate safely. A common approach at the global planning level is to define traversability as a \textit{score} or \textit{cost} and combine it with other costs such as path length, trading off between locomotion stability and navigation time~\cite{fan2021step, frey2023fast, mccrory2023bipedal, yang2021real, castro2023does, wermelinger2016navigation}. While incorporating traversability, the works in~\cite{fan2021step, cai2023probabilistic, cai2024evora} use Conditional Value at Risk (CVaR) to incorporate  environmental uncertainty into the planning for risk-sensitive navigation.

At the local planning level, legged robots have extensively employed MPC to handle complex kinodynamic constraints and generate dynamically feasible motion plans. Depth-integrated MPC incorporates terrain geometry to guide footstep placement on uneven ground~\cite{grandia2023perceptive, acosta2023bipedal}.
Other works~\cite{fan2021step, shamsah2024terrain} incorporate traversability costs into MPC for further refining the path plan from a global planner. However, these approaches typically fix command velocities and overlook high-level selection of velocity commands.

\section{Instability learning for safe bipedal navigation planning}\label{sec: Instability learning for safe bipedal navigation planning}
\subsection{Instability and Fallover Prediction}\label{sec:Instability and Fallover Prediction}
To identify an effective self-supervised signal to train our learning-based traversability estimator specifically for bipedal navigation, we begin by analyzing various locomotion features for their correlations to fallover risk in bipedal locomotion. We collect walking data from Digit in MuJoCo on randomly generated rough terrains with varying geometries, such as slopes, bumps, and surface roughness. Digit is controlled using a terrain-aware ALIP controller \cite{shamsah2024terrain}, commanded with varying linear and angular velocities. The collected data is then parsed into gait cycles, and locomotion features are computed as the Root-Mean-Square (RMS) of raw signals per cycle. Each gait cycle is then labeled with a fallover event: 1 if a fall occurs within two subsequent steps, and 0 otherwise. All candidate features are normalized by their RMS during in-place stepping on flat ground prior to correlation analysis and training.

With this data, we evaluate the correlation between each feature and fallover events via logistic regression. Table~\ref{tab:instab2fallover} summarizes the results using McFadden’s $R^2$~\cite{mcfadden2021quantitative} and AUC-ROC score~\cite{hanley1982meaning}, where $R^2$ around 0.4 and a score close to 1 generally indicates a strong relationship. While features commonly used in grounded mobile robots, such as IMU signals and traction, show acceptable correlations, the body-to-stance-foot angle (BSFA) described in Fig. \ref{fig:overview} leads to better prediction accuracy. This result is reasonable because the bipedal robot counteracts its body perturbation by placing its foot further from the body to increase control authority~\cite{gibson2022terrain}. We refer to this per-cycle RMS as \emph{BSFA instability} $\delta$. Moreover, based on our dataset, we empirically found that BSFA instability is only marginally influenced by slight variations in foot placement, indicating that the metric is robust to small contact-location errors. Based on this finding, we adopt the BSFA instability as the supervisory signal for training our traversability estimation network to predict bipedal instability.

\subsection{Learning Instability}
We aim to predict the BSFA instability from the terrain and the robot state. In this work, we represent the geometrical features of the environment using a 2.5D elevation map \cite{miki2022elevation}. For the robot state, many components such as instant joint values are highly time-varying and hence difficult to consider as decision variables in planning. However, other states, such as velocity action commands, are relatively less time-varying and play a critical role when the learned traversability is later exploited in planning. Hence, we define our instability estimation model as follows:
\begin{align}
    \delta = p_{\theta_{\delta}}(m_{\rm ego}, a), \label{eq:instability}
\end{align}
where $p_{\theta_{\delta}}(\cdot)$ is the target function that maps $m_{\rm ego}$ and $a$ to $\delta$. More specifically, $m_{\rm ego}$ is a 0.64 m $\times$ 0.64 m robot-centric (yaw-aligned) elevation map patch. $a$ is a command vector $(v,w)^T$ with linear velocity $v$ and angular velocity $w$. We model $\delta$ as a Gaussian distribution with mean $\hat{\delta}$ and standard deviation $\sigma$, where $\sigma$ represents the aleatoric uncertainty arising from perception noise and unmodeled effects of instantaneous joint and robot states, and is subsequently used for risk-sensitive planning. Note that external disturbances introduce unmodeled perturbations to the controller, analogous to sensor noise and mapping errors, and are therefore implicitly captured within the aleatoric uncertainty.

\begin{table}[!t]
\vspace{0.1cm}
    \setlength{\tabcolsep}{3pt}
    \renewcommand{\arraystretch}{1.2}
    \centering
    \caption{Comparison of McFadden's $R^2$ and AUC-ROC scores for various locomotion features, evaluating their effectiveness in predicting fallover risk. The highest values in each column are shown in bold.}
    \label{tab:instab2fallover}
    {\small
\begin{tabular}{c|cc}
\hline
 & McFadden's $R^2$
 & AUC-ROC
  \\ \hline
 \begin{tabular}[c]{@{}c@{}}Energy (Control Effort) \cite{gan2022energy}\end{tabular}
 & \begin{tabular}[c]{@{}c@{}} 0.184 \end{tabular}
 & \begin{tabular}[c]{@{}c@{}} 0.784 \end{tabular}\\ \hline
 \begin{tabular}[c]{@{}c@{}}IMU (PSD) \cite{castro2023does}\end{tabular}
 & \begin{tabular}[c]{@{}c@{}} 0.217 \end{tabular}
 & \begin{tabular}[c]{@{}c@{}} 0.837 \end{tabular}\\ \hline
 \begin{tabular}[c]{@{}c@{}}Traction \cite{cai2023probabilistic, frey2023fast}\end{tabular}
 & \begin{tabular}[c]{@{}c@{}} 0.097 \end{tabular}
 & \begin{tabular}[c]{@{}c@{}} 0.770 \end{tabular}\\ \hline
 \begin{tabular}[c]{@{}c@{}}Tangential/Normal Force Ratio\end{tabular}
 & \begin{tabular}[c]{@{}c@{}} 0.016 \end{tabular}
 & \begin{tabular}[c]{@{}c@{}} 0.806 \end{tabular}\\ \hline
 \begin{tabular}[c]{@{}c@{}}Center of Pressure (x-dir)\end{tabular}
 & \begin{tabular}[c]{@{}c@{}} 0.025 \end{tabular}
 & \begin{tabular}[c]{@{}c@{}} 0.859 \end{tabular}\\ \hline
 \begin{tabular}[c]{@{}c@{}}Center of Pressure (y-dir) \end{tabular}
 & \begin{tabular}[c]{@{}c@{}} 0.059 \end{tabular}
 & \begin{tabular}[c]{@{}c@{}} 0.879 \end{tabular}\\ \hline
 Body-to-stance-foot angle (Ours)
 & \begin{tabular}[c]{@{}c@{}} \textbf{0.356} \end{tabular}
 & \begin{tabular}[c]{@{}c@{}} \textbf{0.906} \end{tabular}\\ \hline
\end{tabular}}
\vspace{-1.2\baselineskip}
\end{table}

\begin{figure}
\centering
\includegraphics[width=\linewidth]{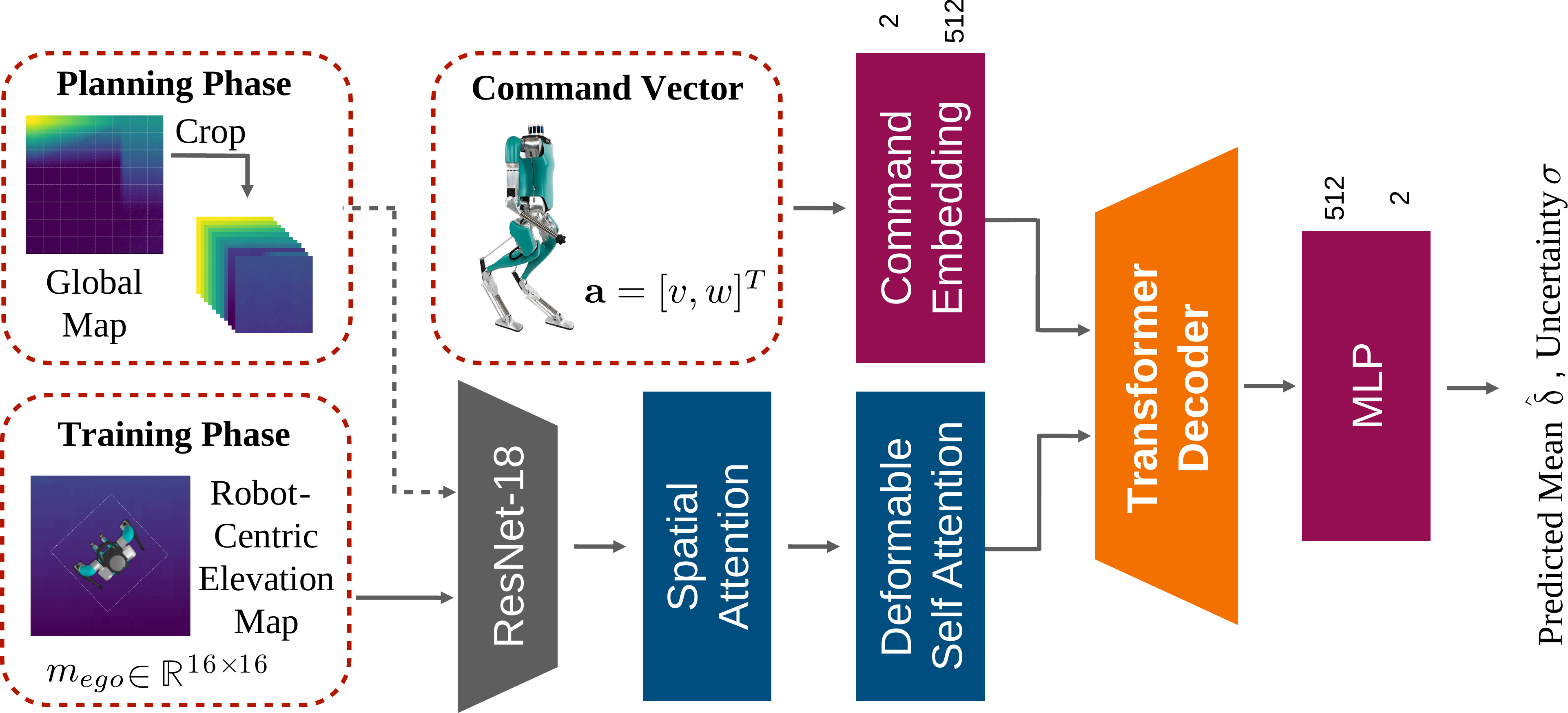}
\caption{TravFormer architecture combines convolutional feature extraction with attention mechanisms to predict BSFA instability. The model processes an elevation map patch $m_{\rm ego}$ representing local terrain and a command vector $a = [v,\omega]^T$. During training, the model uses robot-centric map patches centered on the robot. At planning time, the global terrain map is divided into multiple local patches, which are processed in batch.}
\label{fig:travformer}
  \vspace{-1.5\baselineskip}
\end{figure}

To model this function, we propose \textit{TravFormer}, a neural network architecture designed to predict BSFA instability from the elevation map and command, as shown in Fig. \ref{fig:travformer}. TravFormer leverages a ResNet-18 backbone~\cite{renset} for terrain feature extraction and incorporates attention mechanisms~\cite{attn_is_all_u_need} to focus on stability-critical aspects of the terrain while enabling context-aware integration of the command inputs. The Transformer Decoder treats command embeddings as queries and terrain features as keys/values to generate the final traversability predictions. We use a two-phase schedule for stable joint optimization of $\hat{\delta}$ and $\sigma$, inspired from~\cite{tlio}

\textbf{Phase 1 - Mean Squared Error (MSE) Training:}
In the initial phase, we train the network to minimize the MSE between the network prediction of BSFA instability and the labels collected in Section.~\ref{sec:Instability and Fallover Prediction}.
We train for 10 epochs during this phase, which we find sufficient to establish a strong foundation to ensure convergence for the subsequent uncertainty learning.

\textbf{Phase 2 - Gaussian Negative Log-Likelihood (NLL) Training~\cite{kendall2017uncertainties}}: After initializing with weights from the MSE phase, we extend the network with an additional output head to predict both the mean $\hat{\delta}$ and standard deviation $\sigma$. The network is trained to match the predicted BSFA instability distribution with the label distribution under a Gaussian assumption:
\begin{equation}
\mathcal{L}_{\text{GaussianNLL}} = \frac{1}{N}\sum_{i=1}^{N}\left[\frac{(\hat{\delta}_i - \delta_i)^2}{2\exp(2\log\sigma_i)} + \log\sigma_i\right],
\end{equation}
where $N$ is the total number of samples in a batch and ${\delta}$ denotes the target label collected in Section~\ref{sec:Instability and Fallover Prediction}.

\section{Bipedal Navigation Planning}
\subsection{Traversability as Stability-Aware Command Velocity}\label{subsec: Stability-Aware Command Velocity for Risk-sensitive Navigation}
While prior work often treats traversability as a tunable cost term, such cases are less suitable for bipedal robots, where stability is critical. Instead, we define traversability as the fastest command velocity that satisfies a stability constraint, while accounting for uncertainty in a risk-sensitive manner. 

Given a user-defined instability limit $\delta_{\rm limit}$ and the risk level $\alpha$, we compute the risk-sensitive stability-aware command \mbox{$a^* = (v^*, w^*)$}, which characterizes the traversability of the given local map patch~$m_{\rm ego}$:
\begin{align}
     v^*_{m_{\rm ego}} &= \max \left\{ v \mid \operatorname{VaR}\left( p_{\theta_\delta}(m_{\rm ego}, [v, 0]), \alpha \right) < \delta_{{\rm limit}} \right\}, \label{eq:maximum_v}\\
     \omega^*_{m_{\rm ego}} &= \max \left\{ \omega \mid \operatorname{VaR}\left( p_{\theta_\delta}(m_{\rm ego}, [0, \omega]), \alpha \right) < \delta_{{\rm limit}} \right\}, \label{eq:maximum_w}
\end{align}
\begin{figure}[t]
\centerline{\includegraphics[width=1.\linewidth]{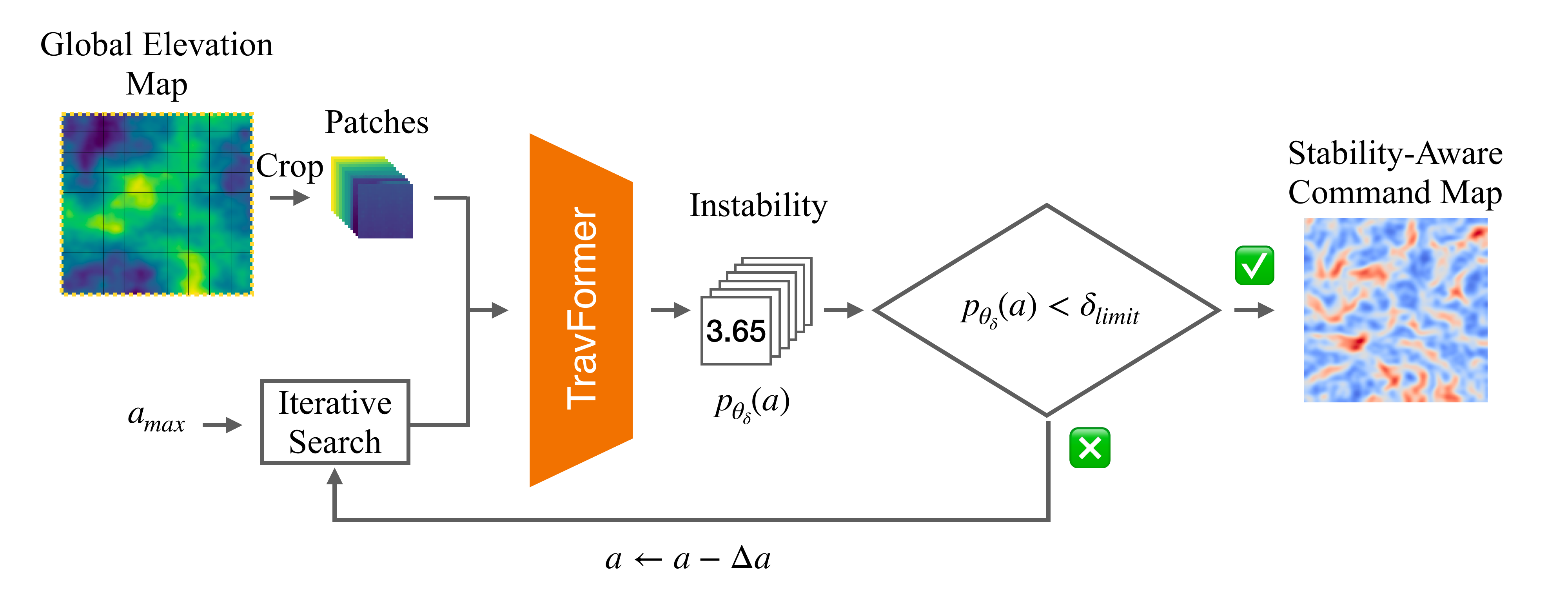}}
\caption{Stability-aware command-velocity selection via iterative search. Starting from the maximum candidate velocity $a_{\max}$, the Iterative Search block feeds $a_{\max}$ into the TravFormer model and then successively replaces it with the next lower candidate sweeping downward until the risk-sensitive stability criterion is met.}
\label{fig:sweeping downward algorithm}
  \vspace{-1.5\baselineskip}
\end{figure}
where $\operatorname{VaR}(\cdot)$ denotes the Value at Risk~\cite{fan2021step}, calculated from the predicted mean and variance under a Gaussian assumption. This fastest velocity is found approximately by sweeping the command velocity from its maximum candidate $a_{\rm max} = (0.5\;m/s, 0.75\;rad/s)$ downward to zero as in Fig. \ref{fig:sweeping downward algorithm}, with a resolution $\Delta a = (0.05\;m/s, 0.075\;rad/s)$. For terrains where the predicted instability exceeds the threshold even at zero command velocity, we assign minimal values of 0.001 m/s and 0.001 rad/s as the stability-aware command velocities, respectively, to penalize traversal through those regions. Note that $\delta_{\rm limit}$ is a robot-specific hyperparameter that does not need to be adjusted according to environmental conditions, but instead serves to configure the robot’s stability during locomotion. In this paper, we use $\delta_{\rm limit} = 3$ and $\alpha = 0.97$. Representing traversability as the fastest command velocity under an instability constraint allows the planner to generate navigation-time-optimal paths without weight tuning, while keeping instability within acceptable bounds. This velocity can also be used to constrain the local MPC planner, ensuring stable execution of the high-level navigation plan.
\begin{rem}[Use of risk-sensitive traversability]
\label{rem:VaR}%
We observe that as command velocity increases, the predicted instability grows in a superlinear manner with increasing uncertainty, particularly on steep terrain. By accounting for this behavior, VaR is incorporated to mitigate overestimation of safe velocities and promote stable navigation under uncertainty.
\end{rem}

\subsection{Traversability-Informed RRT* Global Planner}
We propose traversability-informed RRT* (TravRRT*) built upon the works of~\cite{karaman2011sampling, chen2024adaptive, gammell2014informed} to generate a global path that minimizes navigation time while aiming to keep instability below a user-defined limit and risk level. During sample generation, the traversability-informed RRT* samples nodes with probability proportional to local traversability, encouraging the search toward more navigable paths. For the cost calculation, the cost $c_{ij}$ of an edge connecting a parent node \mbox{$n_i = (x_i,y_i)$} to a child node \mbox{$n_j = (x_j,y_j)$} is defined as the expected traversal time along the edge, subject to stability constraints:
\begin{align}
    c_{ij} &= \sum_{k \in M} \left( \frac{\Delta l_k}{v_k^*} + \frac{\Delta \theta_k}{\omega_k^*} \right),
    \label{eq:edge_cost_proposed}
\end{align}
where $M$ is the set of uniformly sampled intermediate points along the edge. The $\Delta l_k$ and $\Delta \theta_k$ are the linear displacement and heading change at point $k$, while $v^*_k$ and $\omega^*_k$ are the stability-aware linear and angular velocity commands for that point computed by Eq.~\eqref{eq:maximum_v},~\eqref{eq:maximum_w}. By incorporating both translational and rotational traversal time at each intermediate point, this cost formulation allows the planner to select paths that are not only stable but also time-efficient, accounting for both translating and turning motions. 

\subsection{MPC Local Planner}\label{subsec:Local MPC Planner}
To ensure stable execution of the global path generated by the RRT* planner, we formulate an MPC planner that tracks waypoints while accounting for the reduced-order bipedal walking dynamics. We begin with the LIPM to derive discrete dynamics in the local sagittal direction, assuming the fixed step duration \( T \)~\cite{gibson2022terrain}. The CoM position change $\Delta x_q^{\text{loc}}$ between the $q^{\rm th}$ and ${q+1}^{\rm th}$ walking steps, along with the CoM velocity $v_q^{\text{loc}}$, is modeled as a function of the footstep length $u_q^f$ \cite{shamsah2024terrain}.
\begin{align}
\Delta x_q^{\text{loc}}(u_q^f) &= v_q^{\text{loc}} \frac{\sinh(\omega T)}{\omega} + (1 - \cosh(\omega T)) u_q^f , \label{eq:LIP_xcom}\\
v_{q+1}^{\text{loc}}(u_q^f) &= \cosh(\omega T) v_q^{\text{loc}} - \omega \sinh(\omega T) u_q^f,
\label{eq:LIP_xdcom}
\end{align}
where \( \omega = \sqrt{g/H} \) and \( g \) is the gravitational constant and \( H \) is the CoM height. This local dynamics can be transformed into 2D world coordinate by incorporating the robot heading angle $ \phi $ in the world frame:
\begin{align}
    x_{q+1} &= x_q + \Delta x_q^{\text{loc}}(u_q^f) \cos(\phi_q), \label{eq:MPC_xw} \\
    y_{q+1} &= y_q + \Delta x_q^{\text{loc}}(u_q^f) \sin(\phi_q), \label{eq:MPC_yw} \\
    \phi_{q+1} &= \phi_q + u_q^{\Delta \phi},  \label{eq:MPC_phiw}
\end{align}
where \mbox{$u_q^{\Delta \phi}$} is the heading angle change during the step $q$. For better readability, we express the above dynamics \eqref{eq:LIP_xcom}-\eqref{eq:MPC_phiw} compactly as \mbox{$ \mathbf{x}_{q+1} = \Phi(\mathbf{x}_q, \mathbf{u}_q) $} with the state variable \mbox{$\mathbf{x}_q = (x_q, y_q, \phi_q, v^{loc}_q)$}, and the control variable \mbox{$ \mathbf{u}_q = (u_q^f, u_q^{\Delta \phi})$}. Using this dynamics model, we formulate our MPC problem as follows:
\begin{align}
    \min_{\mathbf{x}_q, \mathbf{u}_q } \quad& m(\mathbf{x}_{N+1}) + \sum_{q=0}^{N} l(\mathbf{x}_q, \mathbf{u}_q),\\
    \textrm{s.t.} \quad & \mathbf{x}_0 = \hat{\mathbf{x}}_0, \\
    \quad&\mathbf{x}_{q+1} = \Phi{(\mathbf{x}_q, \mathbf{u}_q)},\\
    \quad& v_{q}^{\text{loc}}/v^* + u_q^{\Delta \phi} / (w^*T) \leq 1,\label{eq:localcmd}
\end{align}
where $\hat{\mathbf{x}}_0$ is the initial state. The cost function includes a terminal cost $m(\mathbf{x}_{N+1})$ which encourages the final heading $\phi_{N+1}$ to align with the direction to the local waypoint $\phi_g$ and the position to reach $(x_g, y_g)$, and a running cost $ l(\mathbf{x}_q, \mathbf{u}_q)$ which penalizes excessive velocities to promote smooth motion:
\begin{align}
    m(\mathbf{x}_{N+1}) &= w_g\Vert (x_{N+1}, y_{N+1}) - (x_g, y_g) \Vert^2 \nonumber\\
    &+ w_\phi\Vert\phi_{N+1}-\phi_g\Vert^2, \\
    l(\mathbf{x}_q, \mathbf{u}_q) &= w_r((v_{q}^{\text{loc}})^2 + (u_q^{\Delta \phi})^2).
\end{align}
Additionally, the stability-aware command velocity imposes a constraint \eqref{eq:localcmd} to regulate command velocities within the stability limit. 

\section{Experimental Results}
\subsection{Learning Instability with TravFormer}
We collected 72{,}000 gait cycles data which corresponds to 8 hours simulation time in total for the network training. For evaluation, we tested three ablations: (i) TravFormer without spatial attention, (ii) TravFormer without deformable attention, and (iii) TravFormer with an transformer decoder replaced by an MLP decoder. We report the root mean squared error (RMSE) and the absolute difference of prediction interval coverage probability ($|\Delta\text{PICP}|$) for all samples, with results shown in Table \ref{tab:ablation_table}. 
\begin{table}[H]
\caption{RMSE and $|\Delta\text{PICP}|$ results on 3 ablated variants and the full version of TravFormer. $|\Delta\text{PICP}|$ represents the absolute difference between the proportion of ground truth instability values fall within one standard deviation of the predicted range (constructed using predicted instability value $\hat{\delta}$ and standard deviation $\sigma$, assuming Gaussian distribution) and the ideal coverage of 68.27\%. The best values are marked in bold.}
\centering
\begin{tabular}{lcc}
\toprule
Variant & RMSE & $|\Delta\text{PICP}|$ \\
\midrule
TravFormer (Full) & 0.7065 & \textbf{0.0007} \\
TravFormer w/o Spatial Attention & 0.7025 & 0.0564 \\
TravFormer w/o Deformable Self-Attention & \textbf{0.6962} & 0.0098 \\
TravFormer w/ MLP Decoder & 0.7958 & 0.1571 \\
\bottomrule
\end{tabular}
\label{tab:ablation_table}
\vspace{-0.1in}
\end{table}
The TravFormer variants with the transformer decoder achieve similarly strong performance, outperforming the MLP decoder by 11.8\% in RMSE for instability prediction. Furthermore, the full model—which combines spatial attention with deformable self-attention—achieves the best uncertainty estimation among all variants.

\begin{figure}[t]
\centerline{\includegraphics[width=1.\linewidth]{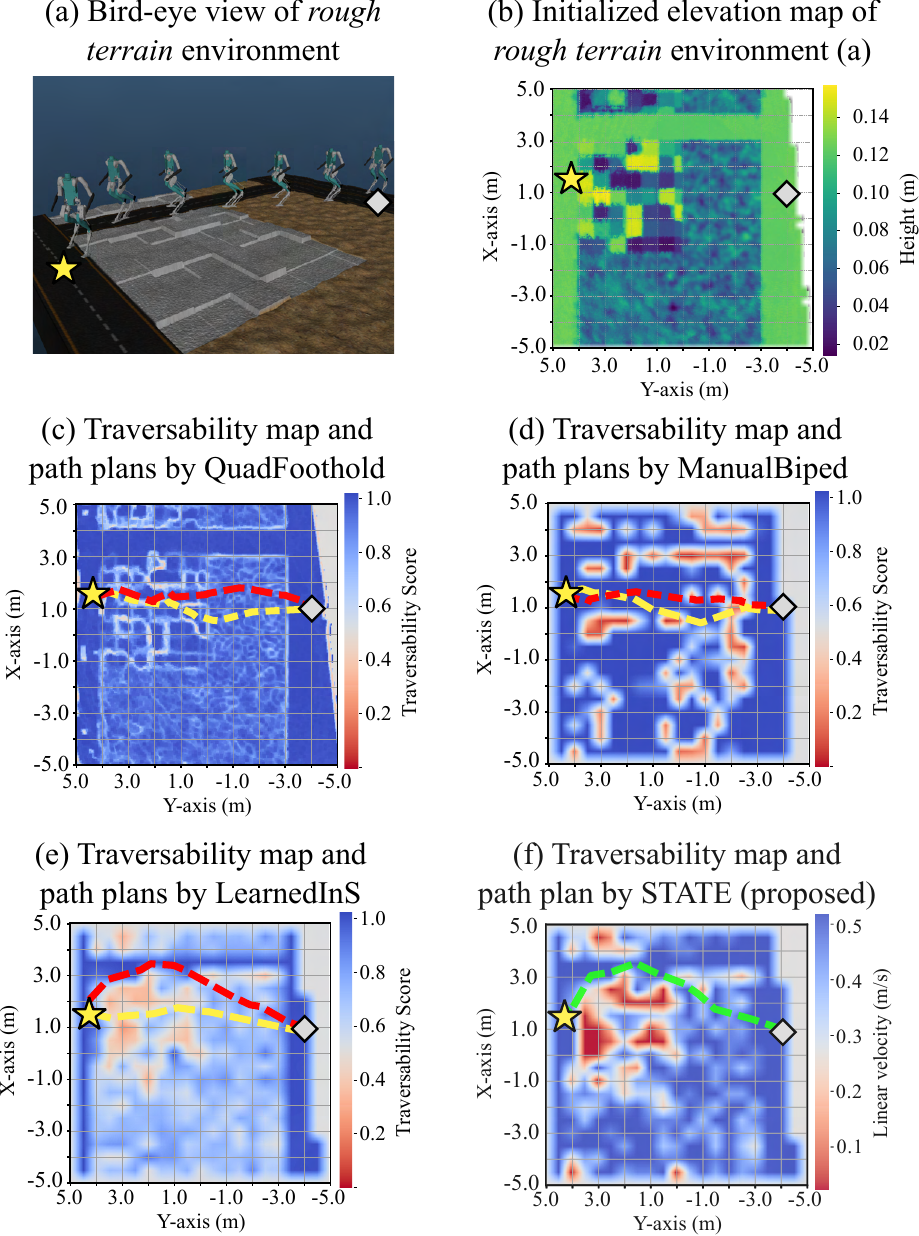}}
\caption{Comparison of estimated traversability maps and paths planned by the proposed method and baselines.
The starting and goal points are marked by a gray diamond and a yellow star, respectively. (a) Diagonal bird-eye view of \textit{rough terrain} environment and the snapshots of the robot traversing along the path by the proposed method. (b) Initialized elevation map of the MuJoCo environment (a). (c), (d), (e) Traversability maps based on (b) and path plans by the QuadFoothold~\cite{wellhausen2021rough},  ManualBiped~\cite{mccrory2023bipedal}, and LearnedInS baselines, respectively. Red and yellow paths are the results of different trade-off weight values $w=3$ and $0.5$. (f) Traversability map based on (b) and the path plan of the proposed method, STATE, with $\delta_{\rm limit} = 3$.
 } 
\label{fig:trav_and_path_plan_site}
  \vspace{-1\baselineskip}
\end{figure}

\begin{figure}[t]
\centerline{\includegraphics[width=0.95\linewidth]{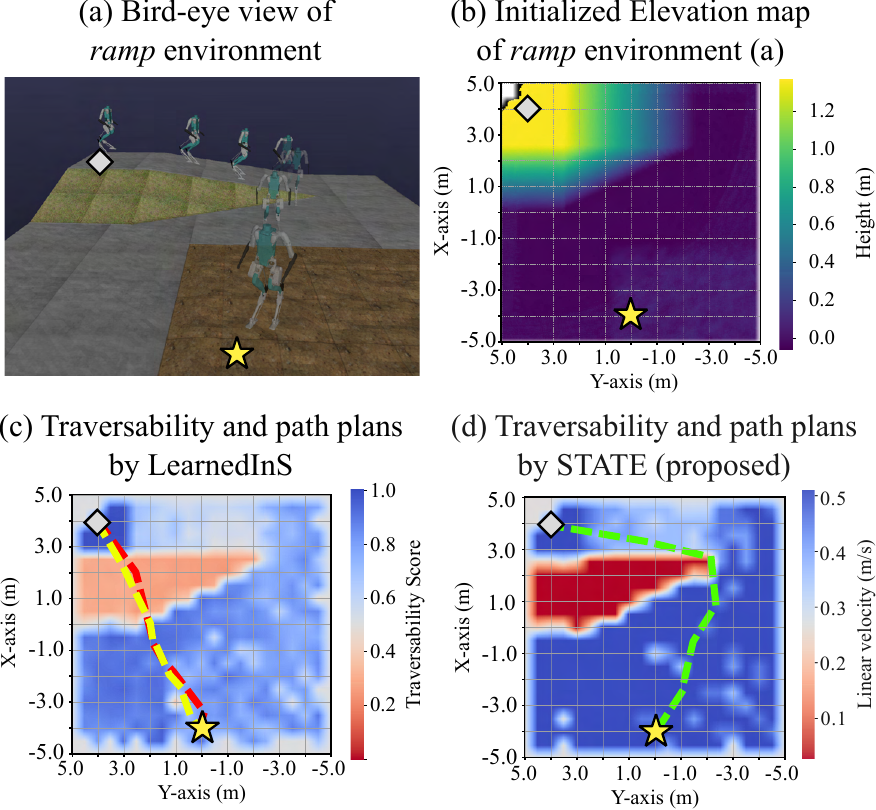}}
\caption{Comparison of estimated traversability maps and paths planned by the proposed method and baselines.
The starting and goal points are marked by a gray diamond and a yellow star, respectively. (a) Diagonal bird-eye view of \textit{ramp} environment and the snapshots of the robot traversing along the path by the proposed method. (b) Initialized elevation map of the MuJoCo environment (a). (c) Traversability map based on (b) and path plans by the LearnedInS baseline. Red and yellow paths are the results of different trade-off weight values $w=3$ and $0.5$. (d) Traversability map based on (b) and the path plan of the proposed method, STATE, with $\delta_{\rm limit} = 3$.
}
\label{fig:trav_and_path_plan_roads}
  \vspace{-1.2\baselineskip}
\end{figure}

\subsection{Experimental Setup}
\subsubsection{Implementation}
We run Digit in MuJoCo simulation with point cloud measurements generated by the simulated sensors. The RRT* has an iteration number of 500, and the MPC planner has a horizon length $N$ of 5 walking steps with \mbox{$w_g = 1$}, \mbox{$w_{\phi} = 5$}, and \mbox{$w_r = 0.1$}. At the beginning of each test, the robot turns around in place at the starting location for 20 seconds to scan the environment and initialize the elevation map. The elevation map and the traversability of the area are kept updated while the robot explores the environment.

\subsubsection{Planner Benchmarking}
To demonstrate the advantages of the proposed method, we design a set of baselines that combine different traversability metrics with various global and local planning strategies. For the cost formulation in global planning, instead of \eqref{eq:edge_cost_proposed}, a trade-off between traversability and path length $ c_{ij} = \sum_{k \in M} \left(1 + w_t(1/t_k) \right) \Delta l_k$ is used for baselines, where $t_k$ is the \textit{traversability score} of each approach, and $w_t$ is the tunable weight \cite{wermelinger2016navigation}. We test two different weight parameters $w = 0.5$ and $3$, to evaluate the effect of the weight on the navigation performance. For local planning, all planners, including the proposed method and the baselines, use the MPC planner introduced in \ref{subsec:Local MPC Planner}. Although the baselines do not define traversability in terms of velocity, we extend their MPC implementations to include a comparable velocity constraint as in~\eqref{eq:localcmd}, ensuring a fair comparison. Specifically, each baseline uses its $v^*, \omega^*$ proportional to its respective traversability score, $t_k a_{\rm max}$, where $a_{\rm max} = (0.5 \;m/s,\; 0.75\;rad/s)$ is the maximum velocity used in Section~\ref{subsec: Stability-Aware Command Velocity for Risk-sensitive Navigation}. 
In summary, we test the planners as follows:
\begin{itemize}
    \item \textbf{Stability-Aware Traversability Estimation (STATE)}: The planner utilizes the stability-aware command velocity \eqref{eq:edge_cost_proposed}, computed with $\delta_{\rm limit} = 3$ and $\alpha = 0.97$. $\delta_{\rm limit} = 3$ is selected since BSFA instability is roughly 2 for stable forward walking at $0.5$ m/s, so $\delta_{\rm limit} = 3$ gives 50\% more stability margin for stable walking at 0.5 m/s.
    \item \textbf{Learned Instability (LearnedInS)}: This method is based on the learned risk-sensitive instability output \eqref{eq:instability}, predicted with a maximum linear command velocity input of $(0.5\;m/s, 0)$ and $\alpha = 0.97$. We take the scaled inverse of the instability to generate a traversability score $t_k$, normalized so that flat terrain has a score of 1, and the higher instability prediction is mapped to a lower score.
    \item \textbf{Manual Traversability Cost Design for Bipeds (ManualBiped) \cite{mccrory2023bipedal}}: This method uses the manual cost design introduced in \cite{mccrory2023bipedal}, which penalizes the slopes and height differences above certain thresholds. Since this design was originally designed to calculate a cost, we apply a scaled inverse transformation in the same manner as in LearnedInS to obtain a traversability score.
    \item \textbf{Quadrupedal Foothold Score (QuadFoothold) \cite{wellhausen2021rough}}: This method is based on the foothold quality score proposed in \cite{wellhausen2021rough}, originally developed for indicating valid area that quadrupeds can reliably take its foothold on.  
\end{itemize}

\subsubsection{Testing Environments}
We validate our approach in various simulated environments. We test the algorithms 10 times each in three 10 m x 10 m MuJoCo environments in Figs. \ref{fig:trav_and_path_plan_site}-
\ref{fig:challenging_terrain}: \textit{rough terrain, ramp, challenging terrain}, respectively. 

\begin{table*}[t]
    \centering
    \renewcommand{\arraystretch}{1.5}
    \caption{Navigation results for \textit{rough terrain} and \textit{ramp} environments. Success rate is recorded by counting the number of trials that succeeded in reaching the goal. Maximum instability values are taken from the instability recording of each successful trial, and then averaged across the trials. Path smoothness is computed as the mean magnitude of angular acceleration within each successful trial, then averaged across the trials, which increases when the robot changes its heading more often. The best values among the planners are marked in bold, and all other values are annotated with the percentage difference relative to the best value.}
    \label{tab:navigation results}
    \resizebox{\textwidth}{!}{%
\begin{tabular}{c|c|cc|cc|cc|cc}
\hline
 \multicolumn{2}{c|}{Traversability Estimation} & \multicolumn{2}{c|}{STATE (proposed)} & \multicolumn{2}{c|}{LearnedInS} & \multicolumn{2}{c|}{ManualBiped} & \multicolumn{2}{c}{QuadFoothold}\\
\cline{1-10}
 \multicolumn{2}{c|}{Global Planner Hyperparameter} & \multicolumn{2}{c|}{$\delta_{\rm limit}=3$} & $w=0.5$ & $w=3$ & $w=0.5$ & $w=3$ & $w=0.5$ & $w=3$ \\
\hline\hline

Rough &Success Rate
& \multicolumn{2}{c|}{\textbf{100\%}}
& 90\%
& \textbf{100\%}
& 90\%
& 90\%
& 80\%
& 70\%
\\ \cline{2-10}


terrain &Max Instability
& \multicolumn{2}{c|}{\textbf{4.82 (+0\%)}}
& 7.90 (+64\%)
& 5.37 (+11\%)
& 7.09 (+47\%)
& 8.31 (+72\%)
& 7.50 (+56\%)
& 7.36 (+53\%)
\\ \cline{2-10}
& Path Smoothness 
& \multicolumn{2}{c|}{\textbf{0.079 (+0\%)}} 
& 0.119 (+50\%) 
& 0.086 (+8.2\%) 
& 0.106 (+34\%) 
& 0.124 (+56\%) 
& 0.114 (+44\%) 
& 0.102 (+29\%) 
\\ \cline{2-10}


&Navigation Time
& \multicolumn{2}{c|}{
\begin{tabular}[c]{@{}c@{}} 32.6 s \\ (+33\%) \end{tabular}
}
& \begin{tabular}[c]{@{}c@{}} 37.3 s \\ (+55\%) \end{tabular}
& \begin{tabular}[c]{@{}c@{}} 36.3 s \\ (+51\%) \end{tabular}
& \begin{tabular}[c]{@{}c@{}} 29.7 s \\ (+23\%) \end{tabular}
& \begin{tabular}[c]{@{}c@{}} 24.3 s \\ (+0.8\%) \end{tabular}
& \begin{tabular}[c]{@{}c@{}} 24.6 s \\ (+2.1\%) \end{tabular}
& \begin{tabular}[c]{@{}c@{}} \textbf{24.1 s} \\ (+0\%) \end{tabular}
\\ \hline\hline

Ramp &Success Rate
& \multicolumn{2}{c|}{\textbf{100\%}}
& 80\%
& 80\%
& 20\%
& 70\%
& 20\%
& 50\%
\\ \cline{2-10}



&Max Instability
& \multicolumn{2}{c|}{\textbf{4.67 (+0\%)}}
& 7.19 (+54\%)
& 7.39 (+58\%)
& 6.51 (+39\%)
& 5.69 (+21\%)
& 6.89 (+47\%)
& 6.61 (+41\%)
\\ \cline{2-10}
& Path Smoothness 
& \multicolumn{2}{c|}{\textbf{0.062 (+0\%)}} 
& 0.150 (+143\%) 
& 0.156 (+152\%) 
& 0.152 (+146\%) 
& 0.115 (+86\%) 
& 0.148 (+139\%) 
& 0.135 (+118\%) 
\\ \cline{2-10}


&Navigation Time
& \multicolumn{2}{c|}{
\begin{tabular}[c]{@{}c@{}} 38.7 s \\ (+34\%) \end{tabular}
}
& \begin{tabular}[c]{@{}c@{}} 40.8 s \\ (+41\%) \end{tabular}
& \begin{tabular}[c]{@{}c@{}} 42.6 s \\ (+47\%) \end{tabular}
& \begin{tabular}[c]{@{}c@{}} 39.1 s \\ (+35\%) \end{tabular}
& \begin{tabular}[c]{@{}c@{}} 35.0 s \\ (+21\%) \end{tabular}
& \begin{tabular}[c]{@{}c@{}} \textbf{28.9 s} \\ \textbf{(+0\%)} \end{tabular}
& \begin{tabular}[c]{@{}c@{}} 29.5 s \\ (+2.1\%) \end{tabular}
\\ \hline\hline

\end{tabular}%
}
\vspace{-0.2in}
\end{table*}




\subsection{Results}
\subsubsection{Benefit of Traversability Learning for Bipedal Locomotion}\label{sec: Traversability tailored for bipedal locomotion}
We first evaluate how well the proposed traversability measures the difficulty in bipedal locomotion in comparison to prior methods. Fig. \ref{fig:trav_and_path_plan_site} illustrates the traversability maps and the path plans generated by each planner in \textit{rough terrain} scenario. We observe that the QuadFoothold baseline primarily detects risk along the edges of the gray stepping stones, but fails to capture the overall difficulty of the region. ManualBiped baseline also underestimates the risk posed by the gray stepped regions, likely due to manually chosen thresholds and hand-crafted rules, which might not be fine-tuned to this type of terrain. Consequently, both planners plan paths that traverse the hazardous areas regardless of the traversability cost weight, which results in lower success rates and higher instability in Table.~\ref{tab:navigation results}. In contrast, our STATE method and the LearnedInS baseline with a high traversability weight leverage our bipedal instability-supervised neural network, assessing stepped areas as more dangerous and successfully avoiding them. These approaches achieve a 100\% success rate and lower instability records, demonstrating the advantage of a traversability estimator tailored to bipedal stability.

\subsubsection{Benefit of Stability-Aware Command Velocity-based Traversability in Global Planning}\label{sec: Benefit of using stability-constrained command velocity-based traversability in global planning}
We then examine the robustness of the proposed global planning strategy by comparing the planning results in different environments. Figs. \ref{fig:trav_and_path_plan_site} and  \ref{fig:trav_and_path_plan_roads} depict the path planned by STATE and LearnedInS on \textit{rough terrain} and \textit{ramp} environments, respectively. We can observe that STATE plans a safe and efficient path in both environments with a single hyperparameter $\delta_{\rm limit}$. 
In contrast, the planning results of LearnedInS are highly dependent on environments. On \textit{rough terrain}, the weight $w=3$ produces a safe and time-efficient path, comparable to that of STATE. However, on \textit{ramp}, the same weight $w=3$ results in unsafe paths with a lower success rate and higher instability, and even a longer navigation time. This is because the length difference between the shortcut and detour paths is greater in \textit{ramp} than that in \textit{rough terrain}, necessitating a higher weight to trigger a detour. However, our proposed stability-aware command is environment-agnostic, allowing the planner to choose paths that maintain a relatively consistent instability level.
\begin{rem}[Environment-agnostic hyperparameter tuning of STATE]
\label{rem:hyperparameter}%
For the baseline traversability score definition $t \in [0,1]$ with the cost $c_{ij} = \sum_{k \in M} (1 + w(1/t_k))^p \Delta l_k$, the hyperparameters $w$ and $p$ must be re-tuned for each environment, since the score and path length have unrelated scales and units. As a result, in \textit{ramp} environments, we report that no safe paths are obtained with any $w$ smaller than 100 under $p = 1$ (not tested beyond $w = 100$), and stable navigation requires the introduction of $p = 2$. However, with $p = 2$, LearnedInS selects an overly conservative path on \textit{rough terrain}, resulting in a longer navigation time of 38.3~s. In contrast, tuning $\delta_{\text{limit}}$ in STATE is intuitive and robot-specific, requiring no iterative tuning across environments. These results highlight the main advantage of STATE: defining traversability by stability-aware velocity enables consistent and environment-agnostic path planning without tedious hyperparameter tuning.
\end{rem}

\begin{figure}[t]
\centerline{\includegraphics[width=.9\linewidth]{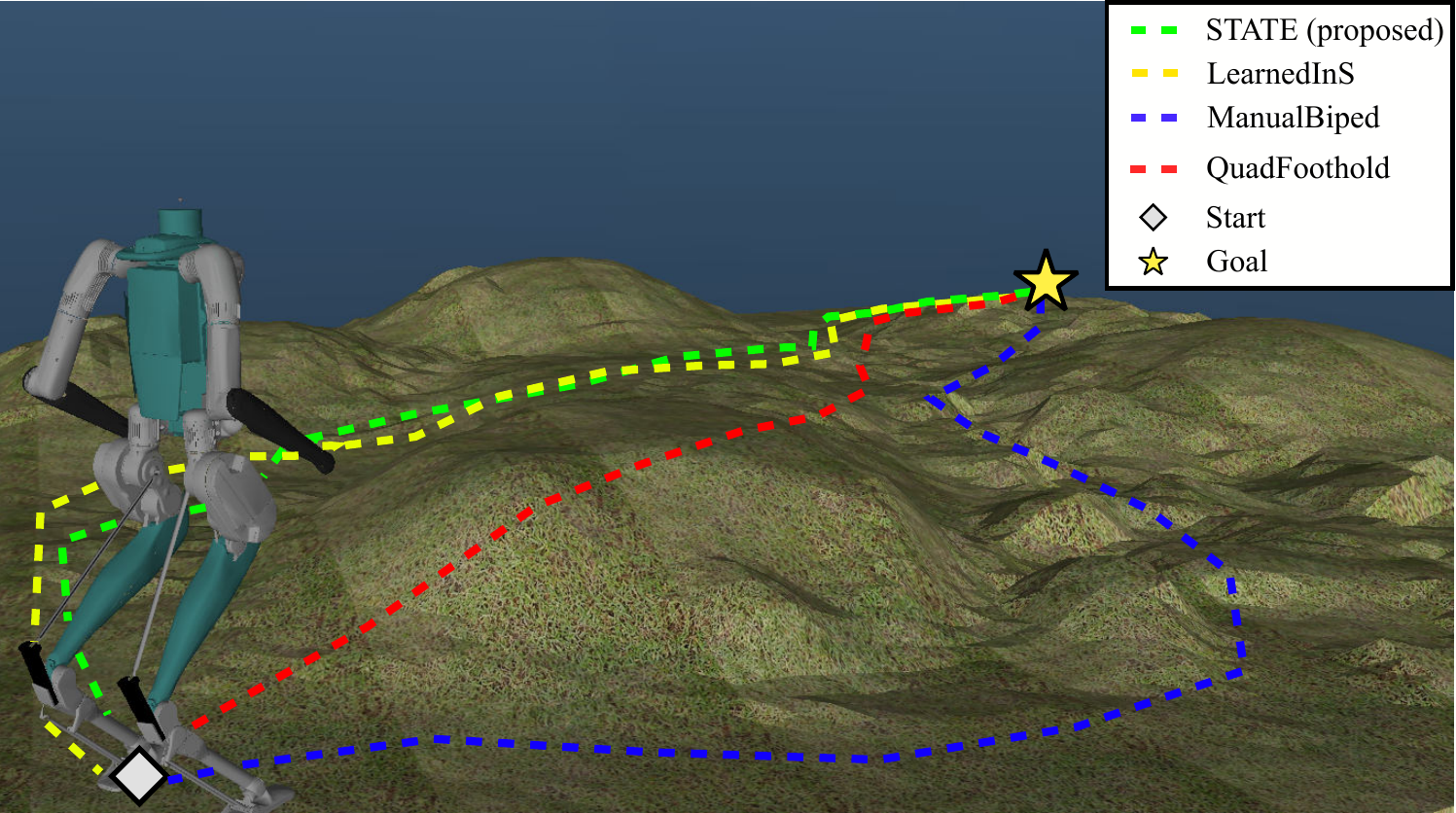}}
\caption{Path plans of the proposed method and the baselines with the weight $w=3$ in \textit{challenging terrain} environment. The starting and goal points are marked by a gray diamond and a yellow star, respectively. Green, yellow, blue, and red paths are by the proposed method (STATE), LearnedInS, ManualBiped, and QuadFoothold planners, respectively.}
\label{fig:challenging_terrain}
  \vspace{-0.5\baselineskip}
\end{figure}

\begin{table}[t]
    \centering
    \renewcommand{\arraystretch}{1.2}
    \caption{Navigation results for \textit{challenging terrain} environment. Evaluation metrics are calculated in the same way as Table.~\ref{tab:navigation results}. 
    }
    \label{tab:navigation results for challenging}
    \setlength{\tabcolsep}{1pt}
    {\small
\begin{tabular}{l|c|c|c|c}
\hline
\multirow{2}{*}{} & \multicolumn{1}{c|}{\begin{tabular}[c]{@{}c@{}} STATE \end{tabular}} & \multicolumn{1}{c|}{\begin{tabular}[c]{@{}c@{}} LearnedInS\end{tabular}} & \multicolumn{1}{c|}{\begin{tabular}[c]{@{}c@{}} ManualBiped \end{tabular} } & \multicolumn{1}{c}{\begin{tabular}[c]{@{}c@{}} QuadFoodhold \end{tabular} }\\
\hline
Success Rate & \textbf{80\%} & {60\%} & {30\%} & 20\% \\ \hline


Max Instability &
\begin{tabular}[c]{@{}c@{}} 6.54 \\ (+3.0\%) \end{tabular} &
\begin{tabular}[c]{@{}c@{}} 7.32 \\ (+15\%) \end{tabular} &
\begin{tabular}[c]{@{}c@{}} \textbf{6.35} \\ \textbf{(+0\%)} \end{tabular} &
\begin{tabular}[c]{@{}c@{}} 12.5 \\ (+97\%) \end{tabular} 
\\ \hline


Path Smoothness &
\begin{tabular}[c]{@{}c@{}}\textbf{0.153}\\\textbf{(+0\%)}\end{tabular} &
\begin{tabular}[c]{@{}c@{}}0.166\\(+8.5\%)\end{tabular} &
\begin{tabular}[c]{@{}c@{}}0.246\\(+61\%)\end{tabular} &
\begin{tabular}[c]{@{}c@{}}0.164\\(+7.2\%)\end{tabular}
\\ \hline


Navigation Time &
\begin{tabular}[c]{@{}c@{}} 61.6 s \\ (+24\%) \end{tabular} &
\begin{tabular}[c]{@{}c@{}} 73.5 s \\ (+48\%) \end{tabular} &
\begin{tabular}[c]{@{}c@{}} 255.3 s \\ (+416\%) \end{tabular} &
\begin{tabular}[c]{@{}c@{}} \textbf{49.5 s} \\ \textbf{(+0\%)} \end{tabular}
\\ \hline
\end{tabular}%
}
  \vspace{-1\baselineskip}
\end{table}

\subsubsection{Benefit of Stability-Aware Command Velocity-based Traversability in MPC Local Planning}\label{sec: Benefit of using stability-constrained command velocity-based traversability in local planning}
Next, we evaluate how stability-aware command velocity improves the execution of global paths in the local MPC planner. Although a global path may be traversable, it often includes regions requiring delicate locomotion, making terrain-aware velocity regulation critical. To evaluate the differences that arise only from local MPC planning behavior, we use \textit{challenging terrain} environment with a baseline weight setting of \mbox{$w=3$}, where both the proposed method and LearnedInS generate nearly identical global paths but the paths contain slopes and rough terrains that demand the robot careful speed regulation, as shown in Fig.~\ref{fig:challenging_terrain}. The navigation results are summarized in Table~\ref{tab:navigation results for challenging}. Note that MPC planners without command constraints, commonly used in prior works, consistently failed and are excluded. The proposed algorithm achieves the highest success rate—though not $100\%$—while maintaining a favorable balance between instability and navigation time shown in Table \ref{tab:navigation results for challenging}. This improvement is attributed to a more effective adaptation of motion aggressiveness to terrain conditions: accelerating in favorable terrain and modulating speed in unstable terrain. By contrast, the linear scaling in baselines cannot fully capture the nonlinear relationship between terrain, command, and locomotion stability. The full simulation results of all four methods are shown in the supplementary video.

\begin{figure}[t]
\centerline{\includegraphics[width=0.95\linewidth]{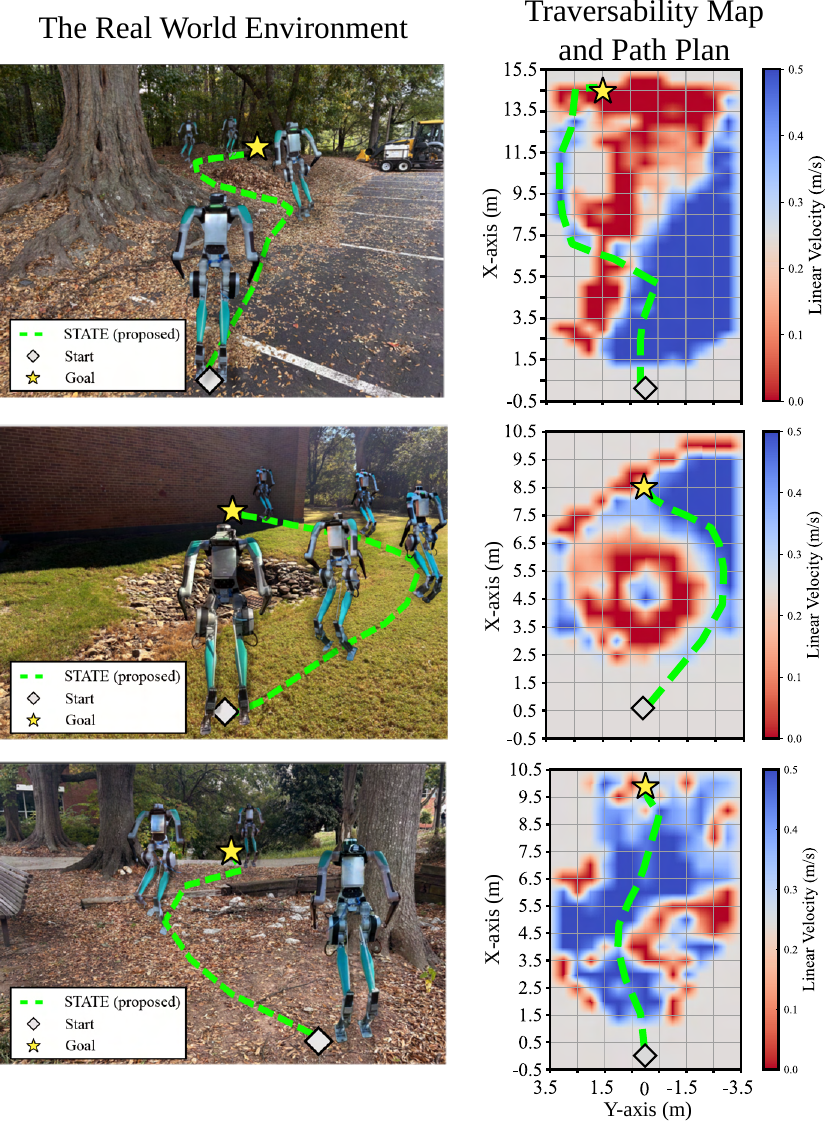}}
\caption{Real-world demonstration of the proposed navigation stack. The left column presents the testing environments with the robot’s executed trajectories overlaid in green. Robot snapshots are extracted from the supplementary video and superimposed for visualization. The right column displays the traversability map and path plan generated by STATE.}
\label{fig:real_world}
  \vspace{-1\baselineskip}
\end{figure}

\subsubsection{Real-World Validation} \label{sec: hardware results}
Beyond simulation, we further validate the proposed framework in real-world settings. The robot perceives its environment using a ZED 2i stereo camera mounted on the chest, providing both point cloud perception and localization. All algorithms are run on an external laptop with an Intel Core i7-12700H CPU and an NVIDIA RTX 3060 Mobile GPU. The framework is transferred to hardware with minimal modifications by increasing the MPC horizon length to $7$ and reducing the number of TravRRT* iterations to $400$. The perception and localization modules operate concurrently, with the traversability map and global plan updated every 5~s, and the MPC local planner running at 3~Hz, ensuring real-time performance. We test the system in three outdoor environments, as shown in Fig.~\ref{fig:real_world}. With this navigation stack, TravRRT* generates safe navigation paths that avoid untraversable regions identified by TravFormer, while the local MPC planner adaptively slows down when crossing ground obstacles or slopes and accelerates on flat terrain to maintain time efficiency. The robot reaches the goal in $54$, $32$, and $34$~s with maximum BSFA instabilities of $3.4$, $2.0$, and $1.7$, respectively, illustrating the practicality of the framework in real world settings. Full experiments, including active velocity modulation, are provided in the supplementary video.


\section{Conclusion and Future Works}
This work presents the first learning-based traversability estimator and navigation framework for bipedal locomotion on diverse rough terrain. Future directions include leveraging the large field of view of humanoids for active exploration of unseen environments. Additionally, incorporating terrain property estimation using semantic perception would be an interesting direction for enhancing robustness.










\let\secfnt\undefined
\newfont{\secfnt}{ptmb8t at 10pt}

\bibliographystyle{IEEEtran}
\bibliography{./bibliography/external_ref}

\end{document}